\documentclass{ws-ijait}

\usepackage{amsmath_jt,multirow}
\usepackage{rotating,tabularx,lscape} 
\begin{document}

\markboth{}
{Kernel Sparse Models for Automated Tumor Segmentation}

%
\catchline{}{}{}{}{}
%

\title{KERNEL SPARSE MODELS FOR AUTOMATED TUMOR SEGMENTATION}

\author{\footnotesize JAYARAMAN J. THIAGARAJAN, KARTHIKEYAN NATESAN RAMAMURTHY, DEEPTA RAJAN and ANDREAS SPANIAS}

\address{School of ECEE, Arizona State University,\\
Tempe, AZ 85287, USA\\
\{jjayaram, knatesan, drajan1, spanias\}@asu.edu}

\author{ANUP PURI and DAVID FRAKES}

\address{School of BHSE, Arizona State University,\\
Tempe, AZ 85287, USA\\
\{anup.puri, dfrakes\}@asu.edu}

\maketitle

\begin{abstract}
\label{sec:abs}
In this paper, we propose sparse coding-based approaches for segmentation of tumor regions from MR images. Sparse coding with data-adapted dictionaries has been successfully employed in several image recovery and vision problems. The proposed approaches obtain sparse codes for each pixel in brain magnetic resonance images considering their intensity values and location information. Since it is trivial to obtain pixel-wise sparse codes, and combining multiple features in the sparse coding setup is not straightforward, we propose to perform sparse coding in a high-dimensional feature space where non-linear similarities can be effectively modeled. We use the training data from expert-segmented images to obtain kernel dictionaries with the kernel K-lines clustering procedure. For a test image, sparse codes are computed with these kernel dictionaries, and they are used to identify the tumor regions. This approach is completely automated, and does not require user intervention to initialize the tumor regions in a test image. Furthermore, a low complexity segmentation approach based on kernel sparse codes, which allows the user to initialize the tumor region, is also presented. Results obtained with both the proposed approaches are validated against manual segmentation by an expert radiologist, and the proposed methods lead to accurate tumor identification.
\end{abstract}


\section{Introduction}
\label{sec:intro}
A robust method to automatically segment a medical image into its constituent heterogeneous regions can be an extremely valuable tool for clinical diagnosis and disease modeling. Given a reasonably large data set, performing manual segmentation is not a practical approach. Brain tumor detection and segmentation have been of interest to researchers recently, however, to this day there exists no comprehensive algorithm built and adopted in the clinical setting \cite{corso2008efficient}. Although patient scans can be obtained using different imaging modalities, Magnetic Resonance Imaging (MRI) has been commonly preferred for brain imaging over other modalities because of its non-invasive and non-ionizing nature, and also because it allows for direct multi-plane imaging.

Tumors may be malignant or benign as determined by a biopsy, and are known to affect brain symmetry and cause damage to the surrounding brain tissues. Automated tumor segmentation approaches are often challenged by the variability in size, shape and location of the tumor, the high degree of similarity in the pixel intensities between normal and abnormal brain tissue regions, and the intensity variations among identical tissues across volumes. As a result, unsupervised thresholding techniques have not been very successful in accurate tumor segmentation \cite{prastawa2003automatic}. Furthermore, approaches that incorporate prior knowledge of the normal brain from atlases require accurate non-rigid registration \cite{kaus2001automated} \cite{clark1998automatic}, and hence generating adequate segmentation results potentially calls for user-intervention and/or a patient specific training system. In addition, these methods require elaborate pre-processing and they tend to over-estimate the tumor volume.

Approaches for tumor segmentation can be either region-based or pixel-based. The active contours method \cite{chan2001active} is a widely adopted region-based approach that is usually combined with a level-set evolution for convergence to a region of interest \cite{ho2002level}. However, it is sensitive to the contour initialization, and has a high computational cost due to its iterative nature. Model-based approaches \cite{moon2002model} employ geometric priors to extend the expectation maximization (EM) algorithm to augment statistical classification. In relatively homogeneous cases such as low grade gliomas, the outlier detection framework proposed by Prastawa \textit{et al.} \cite{prastawa2003automatic} \cite{prastawa2004brain} was shown to perform well.

Pixel-based approaches such as fuzzy C-Means (FCM) using neighborhood labels \cite{ahmed2002modified}, conditional random fields \cite{lee2005segmenting}, Bayesian model-aware affinities extending the SWA algorithm \cite{corso2008efficient}, and the more recent graph-based techniques combined with the cellular-automata (CA) algorithm \cite{hamamci2011tumor} have also achieved some success in tumor segmentation. However, processing issues with respect to contour initialization, noise reduction, intensity standardization, cluster selection, spatial registration, and the need for accurate manual seed-selection leaves substantial room for improvement. In addition, building a robust automated approach that does not require user intervention is very important, particularly for processing large datasets.

\subsection{Sparsity in Tumor Segmentation}
Sparse models form an important component in image understanding since they emulate the activity of neural receptors in the primary visual cortex of the human brain. Olshausen and Field demonstrated that learning sparse linear codes for natural images results in a family of localized, oriented, and bandpass features, similar to those found in the primary visual cortex \cite{Olshausen1997}. Sparsity of the coefficients has been exploited in a variety of signal, and image processing applications including compression \cite{Elad2006}, denoising \cite{elad2006image}, compressed sensing \cite{donoho2006compressed}, source separation \cite{zibulevsky2001blind}, face classification \cite{wright}, and object recognition \cite{lcc}.

Despite its great applicability, the use of sparse models in complex visual recognition applications presents three major challenges: (i) linear generative model of sparse coding can be insufficient for modeling the non-linear relationship between the complex image features, (ii) in several visual recognition tasks, no single descriptor can efficiently model the whole data set, i.e., there is a need to integrate multiple image features into the sparse coding paradigm, and (iii) sparse models require data samples to be represented in the
form of feature vectors, and it is not straightforward to extend them to the case of other forms such as pixel values, matrices or higher order tensors. In order to circumvent the aforementioned challenges, kernel learning methods can be incorporated in sparse coding \cite{JT_MKSR}. Kernel methods map the data samples into a high-dimensional feature space, using a non-linear transformation, in which the relationship between the features can be represented using linear models. Since the resulting feature space is a Hilbert space, kernel methods simplify computations by considering similarities between the features, and not the features themselves. By developing approaches for sparse coding and dictionary learning in the feature space, novel frameworks can be designed for computer vision tasks such as recognition and segmentation. 

In this paper, we develop a novel approach to automatically segment active (enhancing) and necrotic tumor components from T1-weighted contrast-enhanced MR images. We propose to compute kernel sparse codes for the pixels in the image and perform pixel-based segmentation using those codes. Furthermore, we develop the kernel K-lines clustering algorithm to learn kernel dictionaries for coding the pixels. The proposed algorithm for localizing the active tumor regions uses an ensemble kernel constructed using pixel intensities and their spatial locations. Each pixel is classified as belonging to a tumor or a non-tumor region using a linear SVM on the kernel sparse codes. Finally, we propose a semi-automated segmentation technique for improved computational efficiency, wherein the user can initialize the tumor region. This approach eliminates the need to incorporate the spatial location information and also reduces the number of pixels to be processed. In addition, we show that the linear SVM classifier can be replaced by a simple error-based classifier without compromising the segmentation quality. We evaluate the proposed algorithm on a set of T1-weighted contrast-enhanced MR images and compare the results with manual segmentation performed by an expert radiologist. We also show that the proposed algorithms provide accurate segmentation results that outperform the widely used Chan-Vese active contour method \cite{chan2001active}.

\section{Sparse Coding and Dictionary Learning}
\label{sec:sc}

Sparse models have been successful in image understanding because many naturally occurring images can be efficiently modeled as a sparse linear combination of elementary features \cite{JT_MLD}. The elementary features, also referred to as \textit{atoms}, are normalized to unit $\ell_2$ norm and stacked together to form the dictionary matrix. Given a sample $\mathbf{y} \in \mathbb{R}^{M}$, and a dictionary $\mathbf{D} \in \mathbb{R}^{M \times K}$, the generative model for sparse coding is given as $\mathbf{y} = \mathbf{D}\mathbf{x}+\mathbf{n}$, where $\mathbf{x} \in \mathbb{R}^{K}$ is the sparse code with a small number of non-zero coefficients and $\mathbf{n}$ is the noise component. The sparse code can be computed by solving the convex problem
\begin{equation}
\min_\mathbf{x} \|\mathbf{y} - \mathbf{D} \mathbf{x}\|_2^2+ \beta \|\mathbf{x}\|_1,
\label{eqn:sc}
\end{equation} where $\|.\|_1$ indicates the $\ell_1$ norm, and is a convex surrogate for the $\ell_0$ norm which counts the number of non-zero elements in a vector \cite{donoho2003}. Some of the algorithms used to solve (\ref{eqn:sc}) include the basis pursuit \cite{Chen_BP}, feature-sign search \cite{Lee2007} and the least angle regression algorithm with the LASSO modification (LARS-LASSO) \cite{lars}. When presented with a sufficiently large set of training data samples, $\mathbf{Y} = [\mathbf{y}_i]_{i=1}^T$, the dictionary can be learned, and the corresponding sparse codes can be obtained by solving
\begin{equation}
\min_{\mathbf{D},\mathbf{X}} \|\mathbf{Y} - \mathbf{D} \mathbf{X}\|_F^2+ \beta \sum_{i=1}^T\|\mathbf{x}_i\|_1,
\label{eqn:dictlearn_sc}
\end{equation} where $\mathbf{X} = [\mathbf{x}_i]_{i=1}^T$, and $\|.\|_F$ denotes the Frobenius norm of the matrix. Eqn. (\ref{eqn:dictlearn_sc}) can be solved as an alternating minimization problem, where the dictionary is learned by fixing the sparse codes, and the sparse codes are obtained by fixing the dictionary. Dictionaries adapted to the data have been shown to provide superior performance when compared to predefined dictionaries in several applications \cite{Rubin2010} \cite{JT_MLD}. In addition to being useful in data representation problems, there has been a recent surge of interest in using sparse models in several supervised, semi-supervised and unsupervised learning tasks such as clustering \cite{Ramirez} and classification \cite{wright}.

\section{Kernel Sparse Coding for Tumor Segmentation}
\label{sec:ksc}
Sparse coding algorithms are typically employed for vectorized patches or feature vectors extracted from the images, using an overcomplete dictionary. However, the proposed tumor identification algorithm aims to obtain sparse codes for the pixel values directly. This is trivial if we use the approach specified in (\ref{eqn:sc}), since $M=1$ in this case. Furthermore, in order to discriminate between the pixels belonging to multiple segments, we may need to consider the \textit{non-linear similarity} between them. Though the linear generative model of sparse coding has been effective in several image understanding problems, it does not consider the non-linear similarities between the training samples. 

It is typical in machine learning methods to employ the kernel function to learn linear models in a feature space that captures the non-linear similarities. Kernel functions map the non-linear separable features into a feature space $\mathcal{F}$ using a transformation $\Phi(.)$, in which similar features are grouped together. By performing sparse coding in the feature space $\mathcal{F}$, we can obtain highly discriminative codes for samples from different classes \cite{Gao2010}. Note that the choice of the non-linear transformation is crucial to ensure discrimination. The transformation $\Phi(.)$ is chosen such that $\mathcal{F}$ is a Hilbert space with the reproducing kernel $\mathcal{K}(.,.)$ and hence the non-linear similarity between two samples in $\mathcal{F}$ can be measured as $\mathcal{K}(\mathbf{y}_i,\mathbf{y}_j) = \Phi(\mathbf{y}_i)^T\Phi(\mathbf{y}_j)$. Note that the feature space is usually high-dimensional (sometimes infinite) and the closed form expression for the transformation $\Phi(.)$ may be intractable or unknown. Therefore, the \textit{kernel trick} is used to simplify the computations by expressing them in terms of inner products $\Phi(\mathbf{y}_i)^T\Phi(\mathbf{y}_j)$, which can then be replaced using $\mathcal{K}(\mathbf{y}_i,\mathbf{y}_j)$, the value of which is always known. Note that in order for a kernel to be valid, the kernel function or the kernel matrix should be symmetric positive semidefinite according to Mercer's theorem \cite{Cristianini2000}.

\begin{figure}[t]
\begin{minipage}[b]{0.48\linewidth}
  \centering
  \centerline{\psfig{file=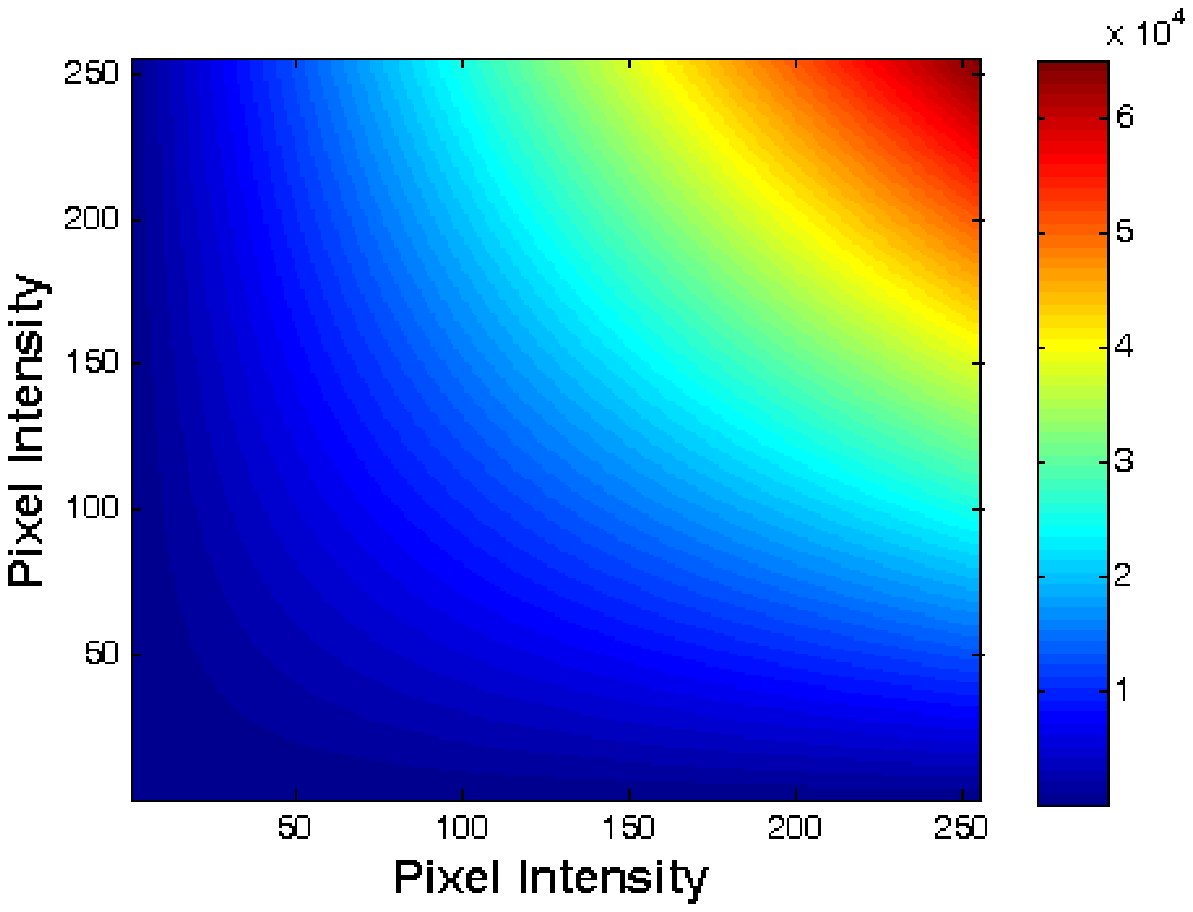,width=6cm}}
  \centerline{(a)}\medskip
\end{minipage}
\hfill
\begin{minipage}[b]{0.48\linewidth}
  \centering
  \centerline{\epsfig{file=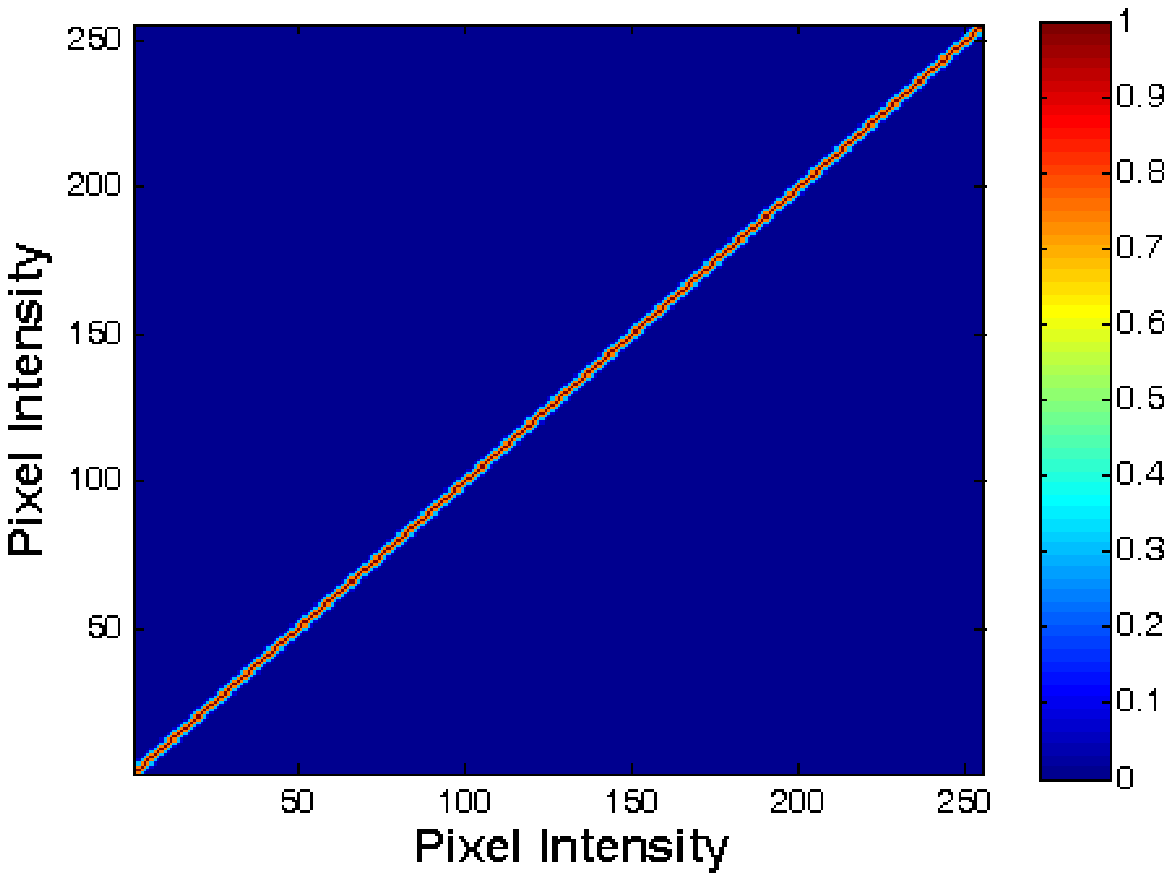,width=6cm}}
  \centerline{(b)}\medskip
\end{minipage}
\caption{Similarity between grayscale pixel intensities ($0$ to $255$): (a) linear similarity ($y_i y_j$) and (b) non-linear similarity ($\mathcal{K}(y_i,y_j)$) using an RBF kernel.}
\label{Fig:lin_nonlim_sim}
\end{figure}

In this paper, we use the Radial Basis Function (RBF) kernel of the form $\mathcal{K}(y_i,y_j) = \exp(-\gamma (y_i-y_j)^2)$, which leads to discriminative sparse codes. As a simple demonstration, the difference between linear similarity of grayscale pixel intensities ($0$ to $255$) and the non-linear similarities obtained using the RBF kernel ($\gamma = 0.3$) is illustrated in Figure \ref{Fig:lin_nonlim_sim}(a) and (b). The linear similarities depend predominantly on the individual intensities of the pixels and not on the closeness of intensities. Whereas, when the RBF kernel is used, the pixel intensities that are close to each other have high non-linear similarity irrespective of the intensities. Pixels with intensities that are far apart have zero non-linear similarity. Therefore, the pixelwise sparse codes that we obtain using such a kernel will behave similarly.

\subsection{Kernel Sparse Coding}
\label{sec:ksparsecodes}
Given the feature mapping function $\Phi: \mathbb{R}^M \mapsto \mathcal{F}$, the generative model in $\mathcal{F}$ for kernel sparse coding is given by $\Phi(\mathbf{y}) = \Phi(\mathbf{D})\mathbf{x}+\mathbf{n}$. We denote the data sample $\mathbf{y}$ in the feature space as $\Phi(\mathbf{y})$ and the dictionary by $\Phi(\mathbf{D}) = [\Phi(\mathbf{d}_1), \Phi(\mathbf{d}_2),...,\Phi(\mathbf{d}_K)]$. The kernel  similarities $\mathcal{K}(\mathbf{y}_i,\mathbf{y}_j) = \Phi(\mathbf{y}_i)^T\Phi(\mathbf{y}_j)$, $\mathcal{K}(\mathbf{d}_k,\mathbf{y}) = \Phi(\mathbf{d}_k)^T\Phi(\mathbf{y})$ and $\mathcal{K}(\mathbf{d}_k,\mathbf{d}_l) = \Phi(\mathbf{d}_k)^T\Phi(\mathbf{d}_l)$ can be computed using pre-defined kernel functions (RBF in our case). All further computations in the feature space will be performed exclusively using kernel similarities. The problem of sparse coding in (\ref{eqn:sc}) can be posed in the feature space as
\begin{equation}
\min_{\mathbf{x}} \|\Phi(\mathbf{y}) - \Phi(\mathbf{D}) \mathbf{x} \|_2^2 + \lambda \|\mathbf{x}\|_1.
\label{eqn:ksc1}
\end{equation} Expanding the objective in (\ref{eqn:ksc1}) we obtain
\begin{align}
\nonumber &\Phi(\mathbf{y})^T\Phi(\mathbf{y}) - 2 \mathbf{x}^T\Phi(\mathbf{D})^T \Phi(\mathbf{y}) + \mathbf{x}^T \Phi(\mathbf{D})^T \Phi(\mathbf{D})\mathbf{x} + \lambda \|\mathbf{x}\|_1, \\
& = \mathbf{K}_{\mathbf{y}\mathbf{y}} - 2 \mathbf{x}^T \mathbf{K}_{\mathbf{D}\mathbf{y}} + \mathbf{x}^T \mathbf{K}_{\mathbf{D} \mathbf{D}} \mathbf{x} + \lambda \|\mathbf{x}\|_1.
\label{eqn:ksc2}
\end{align} Here, $\mathbf{K}_{\mathbf{y}\mathbf{y}}$ is the element $\mathcal{K}(\mathbf{y},\mathbf{y})$, $\mathbf{K}_{\mathbf{D}\mathbf{y}}$ is a $K \times 1$ vector containing the elements $\mathcal{K}(\mathbf{d}_k,\mathbf{y})$, $\forall k = \{1, \ldots, K\}$ and $\mathbf{K}_{\mathbf{D} \mathbf{D}}$ is a $K \times K$ matrix containing the kernel similarities between the dictionary atoms. Clearly, the modified objective function is similar to the sparse coding problem, except for the use of the kernel similarities in place of linear similarities. Hence, the kernel sparse coding problem can be efficiently solved using the feature-sign search algorithm or LARS. However, it is important to note that the computation of kernel matrices incurs additional complexity. Since the dictionary is fixed in (\ref{eqn:ksc2}), $\mathbf{K}_{\mathbf{D} \mathbf{D}}$ is computed only once and the complexity of computing $\mathbf{K}_{\mathbf{D} \mathbf{y}}$ grows as $O(MK)$.

\section{Kernel Dictionary Design}
\label{sec:kklines}
Optimization of dictionaries in the feature space can be carried out by reposing the dictionary learning procedures using only the kernel similarities. Such non-linear dictionaries can be effective in yielding compact representations, when compared to approaches such as the kernel PCA, and in modeling the non-linearity present in the training samples. In this section, we will describe the formulation of a kernel dictionary learning procedure, and demonstrate its effectiveness in representation and discrimination.

The joint problem of dictionary learning and sparse coding in (\ref{eqn:dictlearn_sc}) is a generalization of $1$-D subspace clustering \cite{JT_khyp}. In order to design the dictionary $\Phi(\mathbf{D})$, we will adapt (\ref{eqn:dictlearn_sc}) to the feature space, with the constraint that only one element in the sparse code can be non-zero. This is a special case of the kernel dictionary learning proposed by Nguyen \textit{et. al.} \cite{Nguyen2012}. This procedure is equivalent to the kernel version of K-lines clustering, which attempts to fit $K$ 1-D subspaces to the training data in $\mathcal{F}$ \cite{JT_khyp}. Though sophisticated kernel dictionaries can be designed, employing dictionaries obtained using this simple clustering procedure results in good performance for our tumor segmentation problem. The clustering problem can therefore be posed as
\begin{equation}
\min_{\mathbf{A},\mathbf{X}} \|\Phi(\mathbf{Y}) - \Phi(\mathbf{Y}) \mathbf{A} \mathbf{X}\|_F^2 \text{ such that } \|\mathbf{x}_i\|_0 \leq 1, \forall i.
\label{eqn:kkl_sc}
\end{equation} Each dictionary atom $\Phi(\mathbf{d}_i)$ corresponds to a cluster center and each coefficient vector $\mathbf{x}_i$ encodes the cluster association as well as the weight corresponding to the $i^\text{th}$ pixel. Let us define $K$ membership sets $\{\mathcal{C}_k\}_{k=1}^K$, where $\mathcal{C}_k$ contains the indices of all training vectors that belong to the cluster $k$. The alternating optimization for solving (\ref{eqn:kkl_sc}) consists of two steps: (a) cluster assignment, which involves finding the association and weight of each training vector and hence updating the sets $\{\mathcal{C}_k\}_{k=1}^K$, and (b) cluster update, which involves updating the cluster center by finding the centroid of training vectors corresponding to each set $\mathcal{C}_k$.

In the cluster assignment step, we compute the correlations of a training sample, with the dictionary atoms as $\Phi(y_i)^T \Phi(\mathbf{D}) = \mathbf{K}_{y_i \mathbf{Y}} \mathbf{A}$. If the $k^\text{th}$ dictionary atom results in maximum absolute correlation, the index $i$ is placed in set $\mathcal{C}_k$, and the corresponding non-zero coefficient is the correlation value itself. For the cluster $k$, let $\Phi(\mathbf{Y}_k) = \Phi(\mathbf{Y}) \mathbf{E}_k$  be the set of member vectors and $\mathbf{x}_k^R$ be the row of corresponding non-zero weights. The cluster update involves solving
\begin{equation}
\min_{\mathbf{a}_k} \|\Phi(\mathbf{Y})\mathbf{a}_k\mathbf{x}_k^R - \Phi(\mathbf{Y}) \mathbf{E}_k\|_F^2.
\label{eqn:kkl_update}
\end{equation} Denoting the singular value decomposition of 
\begin{equation}
\Phi(\mathbf{Y}_k) = \mathbf{U}_k \mathbf{\Sigma}_k \mathbf{V}_k^T,
\label{eqn:svd}
\end{equation} the rank-1 approximation, which also results in the optimal solution for (\ref{eqn:kkl_update}), is given by
\begin{equation}
\Phi(\mathbf{Y})\mathbf{a}_k\mathbf{x}_k^R  = \mathbf{u}_{k_1} {\sigma}_{k_1} \mathbf{v}_{k_1}^T,
\label{eqn:rank1approx}
\end{equation} where ${\sigma}_{k_1}$ is the largest singular value, and $\mathbf{u}_{k_1}$ and $\mathbf{v}_{k_1}$
are the columns of $\mathbf{U}_k$ and $\mathbf{V}_k$ corresponding to that singular value. Eqn. (\ref{eqn:rank1approx}) implies that $\Phi(\mathbf{Y})\mathbf{a}_k = \mathbf{u}_{k_1}$ and $\mathbf{x}_k^R  = {\sigma}_{k_1} \mathbf{v}_{k_1}^T$. Let the eigen decomposition of $\mathbf{K}_{\mathbf{Y}_k \mathbf{Y}_k}$ be $\mathbf{V}_k \mathbf{\Delta}_k \mathbf{V}_k^T$ and hence we have ${\sigma}_{k_1} = \sqrt{\Delta_k(1,1)}$, assuming the eigen values are in descending order. From  (\ref{eqn:svd}), we also have $\Phi(\mathbf{Y}_k)\mathbf{v}_{k_1} = \sigma_{k_1} \mathbf{u}_{k_1}$. Substituting for $\Phi(\mathbf{Y}_k)$ and 
$\mathbf{u}_{k_1}$, we obtain $\Phi(\mathbf{Y}) \mathbf{E}_k \mathbf{v}_{k_1} = \sigma_{k_1} \Phi(\mathbf{Y}) \mathbf{a}_k$, which results in
\begin{equation}
\mathbf{a}_k  = {\sigma}_{k_1}^{-1} \mathbf{E}_{k}\mathbf{v}_{k_1}.
\label{eqn:ak_update}
\end{equation} Note that $\mathbf{a}_k$ completely defines $\mathbf{d}_k$. The cluster assignment and update steps are repeated until convergence, i.e., when $\{\mathcal{C}_k\}_{k=1}^K$ does not change over iterations.

\begin{figure}[t]
\begin{minipage}[c]{0.48\linewidth}
  \centering
  \centerline{\psfig{file=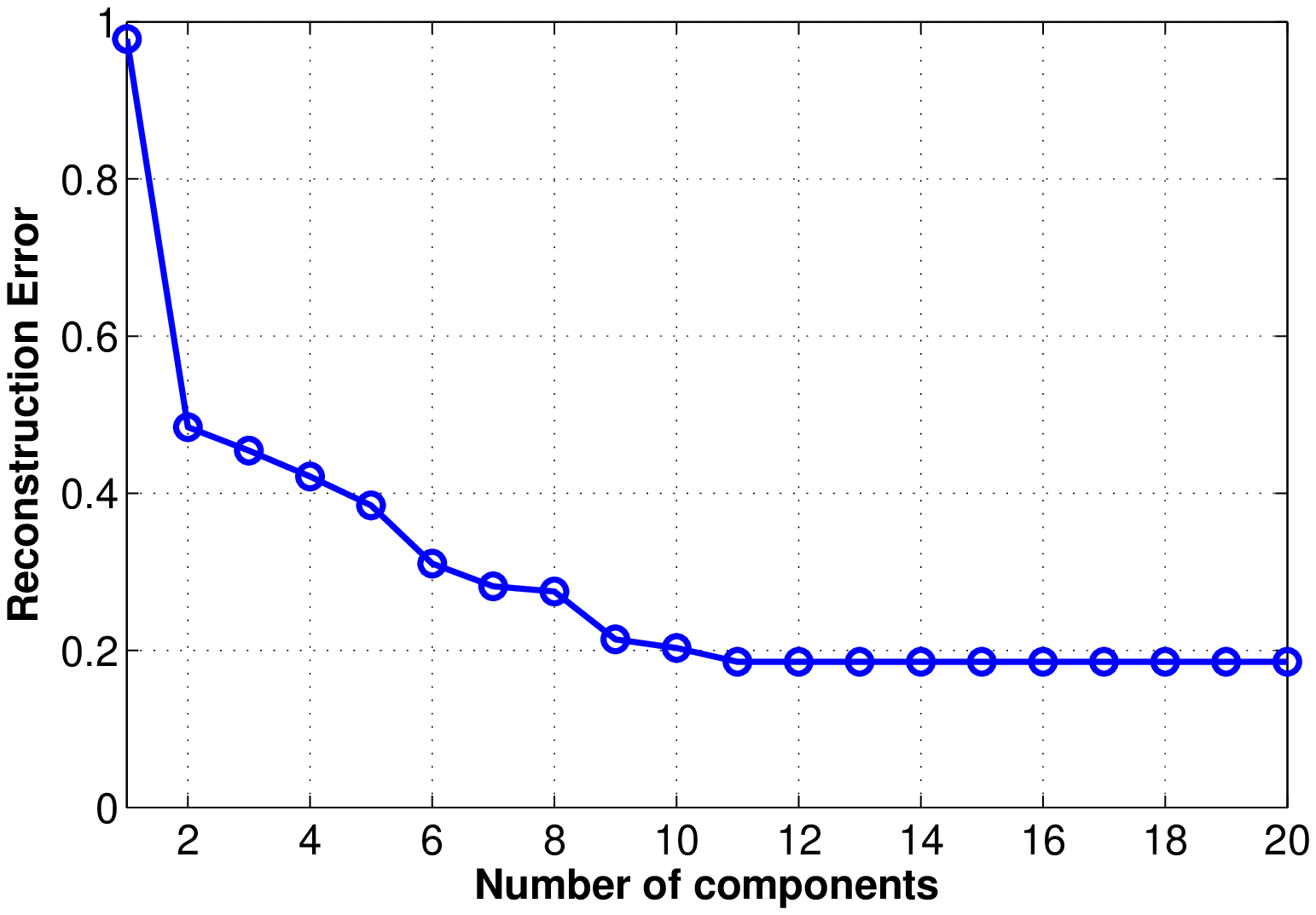,width=7.5cm}}
\end{minipage}
\hfill
\begin{minipage}[c]{0.48\linewidth}
  \centering
  \centerline{\psfig{file=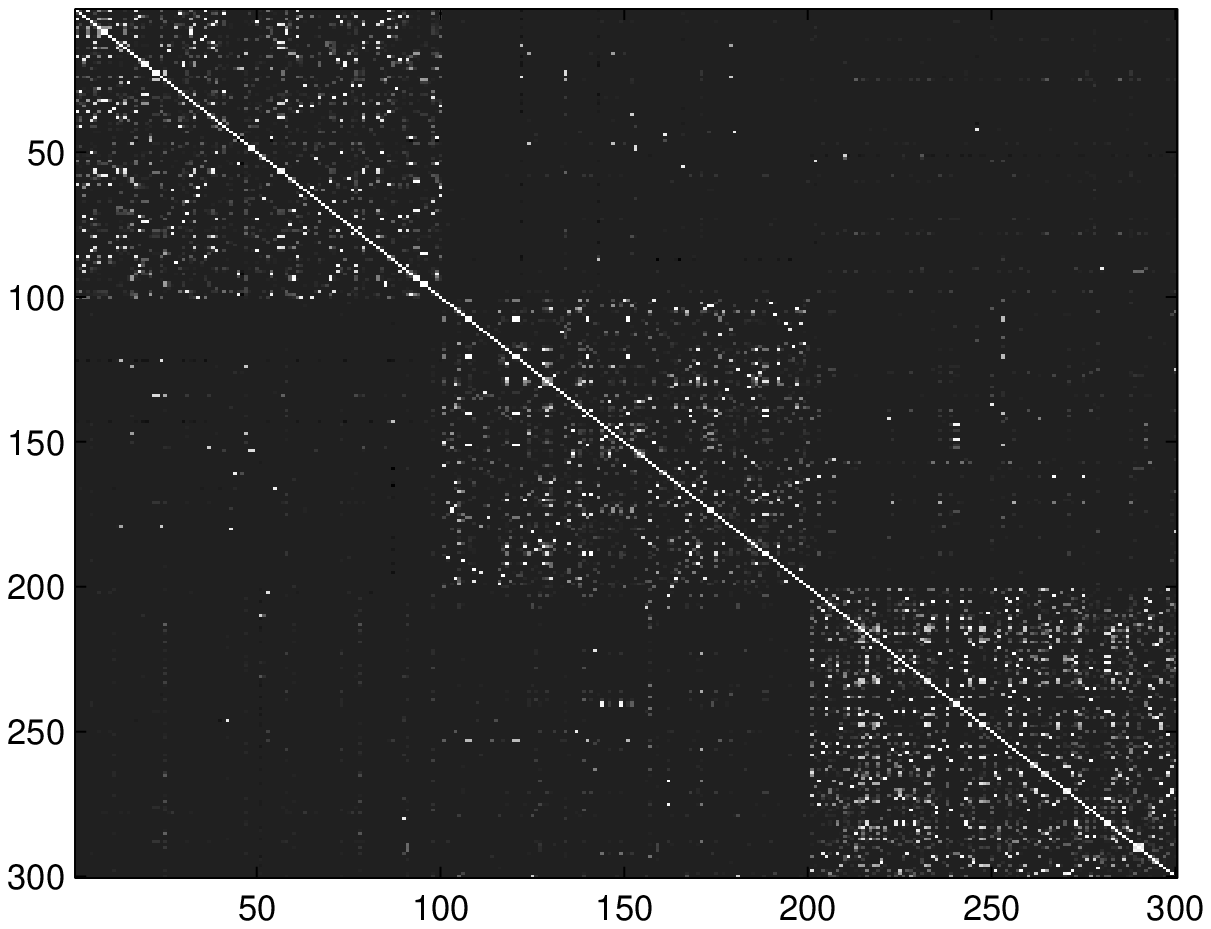,width=6.8cm}}
\end{minipage}
\caption{(a) Reconstruction error for a novel test sample using kernel sparse coding, for different values of sparsity. (b) Similarity between the kernel sparse codes of samples drawn from $3$ different classes in the USPS dataset. Since the kernel codes of samples belonging to the same class are highly similar, we observe a block-wise structure in the normalized correlation plot.}
\label{Fig:ksc_rep}
\end{figure} 

\subsection{Representation}
Kernel sparse coding can be used as an alternative to approaches such as kernel PCA for efficient data representation. Though complete reconstruction of the underlying data from the kernel sparse codes requires computation of pre-images \cite{kwok2004}, novel test samples can be well approximated using the learned kernel dictionaries. As a demonstration, we consider the class of digit $2$ from the USPS dataset \cite{usps} and use a subset of images for training a kernel dictionary using kernel K-lines clustering. For a novel test sample $\mathbf{z}$, different from the training set, we compute sparse code using (\ref{eqn:ksc1}) and compute the reconstruction error as
$\|\Phi(\mathbf{z}) - \Phi(\mathbf{D}) \mathbf{a}\|_2^2$. Figure \ref{Fig:ksc_rep}(a) shows the reconstruction error obtained for a test sample for different number of non-zero coefficients, $\{1, \ldots, 20\}$.

\subsection{Discrimination}
In addition to efficiently modeling data samples, kernel sparse coding is well suited for supervised learning tasks. Since the non-linear similarities between the training samples are considered while learning the dictionary, the resulting codes are highly discriminative. As a demonstration, we consider $100$ training samples each from $3$ different classes in the USPS dataset (Digits $3, 4$ and $7$). We obtain the kernel sparse codes for all the samples and compute the normalized cross correlation between the sparse features. Since kernel sparse codes promote discrimination, we expect the features belonging to a class to be highly similar to each other compared to samples from other classes. The block-wise structure in the normalized correlation plot in Figure \ref{Fig:ksc_rep}(b) evidences this.

\section{Proposed Automated Tumor Segmentation Algorithm}
\label{sec:algo}
Adaptive thresholding and unsupervised segmentation approaches are basic pixel-based approaches for obtaining tumor regions from MR images. However, building more sophisticated tools, by incorporating expert knowledge, can improve segmentation performance. In this section, we describe the proposed algorithm for automated tumor segmentation based on kernel sparse codes. 

To perform tumor segmentation, we need to identify pixels that can possibly constitute a tumor region based on intensity. Though segmentation is as an unsupervised learning problem, we can pose it is as a supervised learning problem since we can easily obtain a at least a few training images with tumor regions marked by an expert. Hence, we propose to obtain kernel dictionaries using the training samples and learn a $2$-class classifier (Tumor vs Non-tumor). Furthermore, in order to localize the tumor regions in the image, we need to incorporate additional constraints to ensure connectedness among pixels in a segment. This can be addressed by building a spatial location kernel and fusing it with the intensity kernel.

\subsection{Combining Multiple Features}
When compared to using a single feature, using multiple features to characterize images has been a very successful approach for several classification tasks. Though this method provides the flexibility of choosing features to describe different aspects of the underlying data, the resulting representations are high-dimensional and the descriptors can be very diverse. Hence, there is a need to transform the features to a unified space that facilitates the recognition tasks, and construct low dimensional compact representations for the images in the unified space.

Let us assume that a set of $R$ diverse descriptors are extracted from a given image. Since the kernel similarities can be used to fuse the multiple descriptors, we need to build the base kernel matrix for each descriptor. Given a suitable distance function $d_r$, which measures the distance between two samples for the feature $r$, we can construct the kernel matrix as
\begin{equation}
\mathbf{K}_r(i,j) = \mathcal{K}_r(\mathbf{y}_i,\mathbf{y}_j) = \exp(-\gamma d_r^2(\mathbf{y}_i,\mathbf{y}_j)), 
\label{eqn:mksc3}
\end{equation}where $\gamma$ is a positive constant. Given the $R$ base kernel matrices, $\{\mathbf{K}_r\}_{r=1}^R$, we can construct the ensemble kernel matrix as
\begin{align}
\mathbf{K} &= \sum_{r=1}^R \beta_r \mathbf{K}_r, \quad \forall \beta_r \geq 0.
\label{eqn:mksc2}
\end{align} A useful alternate approach to fuse the descriptors is to obtain the ensemble kernel matrix as
\begin{align}
\mathbf{K}  = \mathbf{K}_1 \odot \mathbf{K}_2 \odot \ldots \odot \mathbf{K}_R,
\label{eqn:mksc4}
\end{align}where $\odot$ denotes the Hadamard product between two matrices. Sparse codes computed with the ensemble kernel matrices will take all the $R$ features into account. Note that when combining kernel matrices we need to ensure that the resulting kernel matrix also satisfies the Mercer's conditions. 

\begin{figure}[t]
\begin{minipage}[b]{1\linewidth}
  \centering
  \centerline{\psfig{file=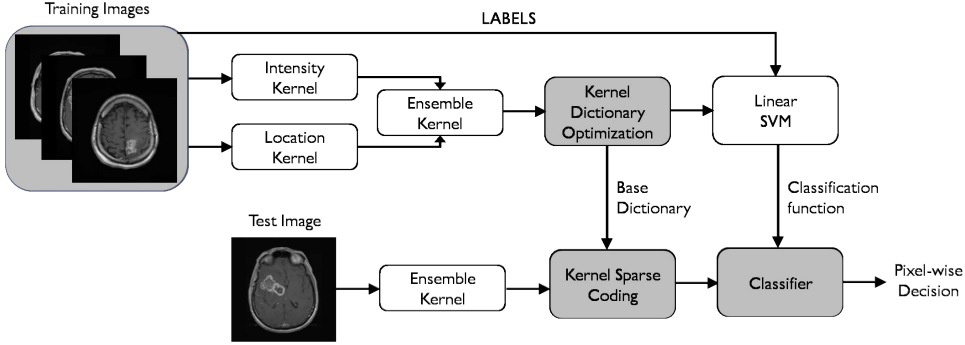,width=13cm}}
 \end{minipage}
\caption{Illustration of the proposed algorithm for automated tumor segmentation. For a set of training samples, the ensemble kernel dictionary is obtained using Kernel K-lines clustering procedure, and a $2$-class linear SVM is used to classify the pixels.}
\label{Fig:auto}
\end{figure}

\subsection{Algorithm}
The proposed algorithm for automated tumor segmentation is illustrated in Figure \ref{Fig:auto}. In the rest of this paper, we refer to this as the \textit{Kernel Sparse Coding-based Automated} (KSCA) segmentation algorithm. In the training stage, it is assumed that  the location of the tumor pixels are known in the ground truth training images. For a subset of $T$ pixels (both positive and negative examples) obtained from the training images, we compute the intensity kernel matrix, $\mathbf{K}_I \in \mathbb{R}^{T \times T}$, by employing an RBF kernel on the pixel intensity values. In addition, the spatial location kernel matrix $\mathbf{K}_L$ is constructed as 
\begin{equation}
\mathbf{K}_L(i,j) = \mathcal{K}_L(y_i,y_j) = \begin{cases} \exp^{\|\mathbf{l}_i - \mathbf{l}_j\|_2^2}, &\mbox{\text{if } } j \in \mathcal{N}(i), \\
0, &\mbox{\text{otherwise.}} \end{cases}
\label{eqn:lknl}
\end{equation}Here, $\mathcal{N}(i)$ denotes the neighborhood of the pixel $y_i$, and $\mathbf{l}_i$ and $\mathbf{l}_j$ are the respective location vectors for the pixels $y_i$ and $y_j$. We fuse the intensity and spatial location kernel matrices to obtain the ensemble kernel matrix,  $\mathbf{K} = \mathbf{K}_I \odot \mathbf{K}_L$. 

The sparse codes obtained with a dictionary learned in the ensemble feature space model the similarities of pixels with respect to both intensity and location of pixels. A set of training images, with active tumor regions, are used to learn a kernel dictionary with the kernel K-lines clustering procedure. Using the kernel sparse codes belonging to tumor and non-tumor regions, we learn a two-class linear SVM to classify the pixel. For a test image, we obtain the required ensemble kernel matrices and compute the kernel sparse codes using the learned dictionary. Finally, the SVM classifier can be used to identify the pixels belonging to an active tumor region. The impact of combining diverse features using kernel sparse coding is evidenced by the accurate segmentation results. 

\begin{figure}[t]
\begin{minipage}[b]{1\linewidth}
  \centering
  \centerline{\psfig{file=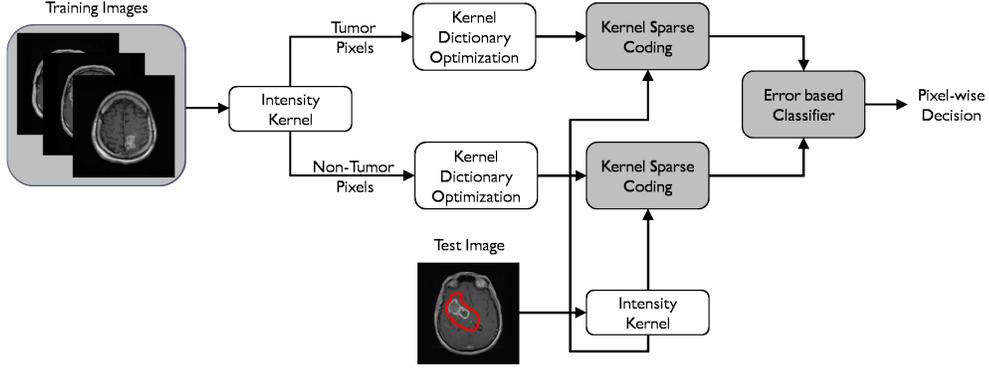,width=13cm}}
 \end{minipage}
\caption{Illustration of the semi-automated approach for complexity reduction in the proposed algorithm. By allowing the user to initialize the tumor region in a test image, the need for incorporating locality information is eliminated. Furthermore, the SVM classifier can be replaced by a simple reconstruction error-based classifier.}
\label{Fig:semi}
\end{figure}

\section{Complexity Reduction using a Semi-Automated Approach}
\label{sec:semi}
The amount of training required and the computational complexity are two important factors that can determine the efficiency of an automated segmentation algorithm. Since the dictionary training is performed using pixels, the number of training images required is quite limited. Though the computational complexity of the automated segmentation algorithm described earlier is comparable to several existing methods, its efficiency can be further improved by allowing the user to initialize the tumor region. Computing the kernel sparse codes for all pixels in a test image incurs the maximum complexity and hence initializing the tumor regions drastically reduces the number of pixels to be processed. Furthermore, there is no need to explicitly include the location information in the algorithm, since the tumor region has already been localized by the user. Hence, the classification can be carried out by using a simple error-based classifier on the kernel sparse codes. We refer to this as the \textit{Kernel Sparse Coding-based Semi-Automated} (KSCSA) segmentation approach. We observed from our experiments that for an average sized tumor region, we achieve significant speedup by using the semi-automated approach. Furthermore, the segmentations obtained using the two methods are quite comparable, though the automated approach can potentially generate more false positives when compared to the semi-automated approach.

Given a set of training images containing active tumor regions, we use the tumor and non-tumor pixels to train two separate kernel dictionaries. We construct two RBF kernel matrices on the pixel intensities and employ the kernel K-lines clustering algorithm to learn the tumor and non-tumor dictionaries, $\Phi(\mathbf{D}_T)$ and $\Phi(\mathbf{D}_N)$, respectively. Note that dictionary learning is performed only once, and as we will show in our experimental results, the dictionaries generalize well to reasonably large datasets.

For a test image, we obtain kernel sparse codes for each pixel $y_i$ using $\Phi(\mathbf{D}_T)$ and $\Phi(\mathbf{D}_N)$, and denote the respective sparse codes as $\mathbf{x}_i^T$ and $\mathbf{x}_i^N$. Since the dictionaries are optimized for two different classes of pixel intensities, we expect the tumor pixels to be better modeled by the tumor dictionary. Hence we classify a pixel as belonging to an active tumor region if the approximation error obtained with the tumor dictionary is less than that obtained with the non-tumor dictionary:
\begin{equation}
\mathcal{J}(y_i) = \begin{cases} \textit{Tumor}, &\mbox{if } E_N - E_T  \geq \epsilon, \\
\textit{Non-tumor}, &\mbox{otherwise.} \end{cases}
\label{eqn:seg}
\end{equation}Here the approximation errors with respect to the two dictionaries are $E_N = \|\Phi(y_i) - \Phi(\mathbf{D}_N) \mathbf{x}_i^N\|_2$ and $E_T = \|\Phi(y_i) - \Phi(\mathbf{D}_T) \mathbf{x}_i^T\|_2$, respectively. Note that the threshold for the error difference, $\epsilon$, can be tuned using a validation dataset before applying the algorithm to the test data. 

\section{Experiments}
\label{sec:exp}

In this section, we provide details about the datasets used to evaluate our algorithm and present the segmentation results. The results are compared to manual segmentations performed by a radio-oncology specialist, based on both the subjective visual quality and quantitative standards such as \textit{Accuracy (Acc)} and \textit{Correspondence Ratio (CR)}. 

\subsection{Dataset}
\label{sec:data}
The algorithm was tested on a set of T1-weighted (spin echo) contrast-enhanced, 2-D Dicom format images acquired with a 1.5T GE Genesis Signa MR scanner. Each axial slice was 5 mm thick with a 7.5 mm gap between slices, and the size of the image matrix was $256 \times 256$. Patients were administered a $20$cc Bolus of Gadolinum contrast agent, and were already diagnosed with Glioblastoma Multiforme (GBM), the most common form of dangerous and malignant primary brain tumor. These tumors are characterized by jagged boundaries with a ring enhancement, possibly a dark core necrotic component, and are accompanied by edema (swelling). The ground truth (GT) images were obtained from the manual segmentation carried out by an expert radiologist at the St. Joseph's Hospital and Medical Center in Phoenix, AZ, USA. We tested our algorithm on the pre- and post-treatment images for $9$ patients where all the slices (approximately $175$) showed the presence of GBM.

\subsection{Benchmark Algorithm - Active Contour Method}
We compare the segmentation results of our proposed algorithms to the widely used Chan-Vese active contour method (ACM) \cite{chan2001active}. The main goal of this region based method is to minimize the energy function defined by the means of the pixel intensities inside and outside the initial level set curve. Note that this algorithm is not completely automated. The initial level set formulation is conveyed to the algorithm by enabling the user to draw a binary mask over the region of interest in the image. The binary mask is converted to a Signed Distance Function (SDF), such that the region within the curve is assigned positive values, increasing with distance, and the region outside the curve is given increasing negative values, with the distance from the curve. The SDF enables interaction with the energy function as it associates the modification and movement of the initial level set formulation with the change in energy statistics in the two regions. An update occurs with every iteration, wherein the curve evolves and a new SDF is generated based on the previous iteration.  The algorithm stops updating the initial level set formulation when the energy is minimized, and further evolution of the curve leads to an increase in the energy value achieved in the previous iteration.

Since this algorithm is not based on gradient methods, and deals with balancing the energy on both sides of the curve, it achieves good results even when the image is blurred. One of the main advantages of this algorithm is that it relies on global properties rather than just taking into account local properties, such as gradients. Furthermore, it provides improved robustness in the presence of noise. 

\subsection{Results}
\label{sec:results}
Simulations were carried out independently for both the semi-automated and automated algorithms for every axial slice. For both of the proposed algorithms, the parameter $\gamma$ for the RBF kernel was set to $0.3$, and the dictionary size was fixed at $256$. In the automated approach, we computed the ensemble kernel for $15,000$ randomly chosen pixels from the training set. In the reduced complexity semi-automated case, the tumor and non-tumor dictionaries were learned using $10,000$ randomly chosen pixels from tumor and non-tumor regions respectively. The parameter $\beta = 0.1$ was used for sparse coding using the feature sign search algorithm. 

\begin{figure}[t]
\begin{minipage}[b]{0.48\linewidth}
  \centering
  \centerline{\psfig{file=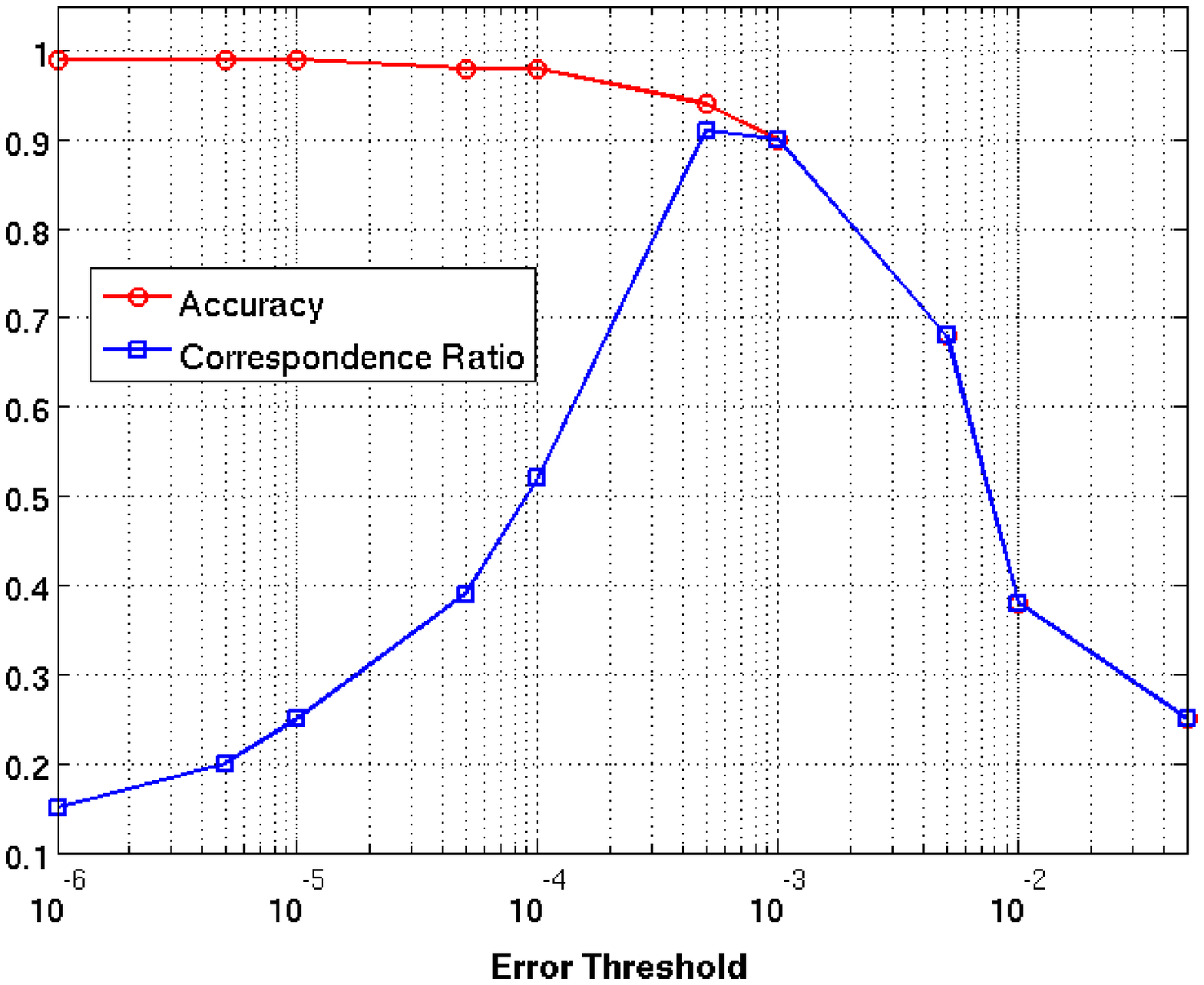,width=6cm}}
\end{minipage}
\hfill
\begin{minipage}[b]{0.48\linewidth}
  \centering
  \centerline{\epsfig{file=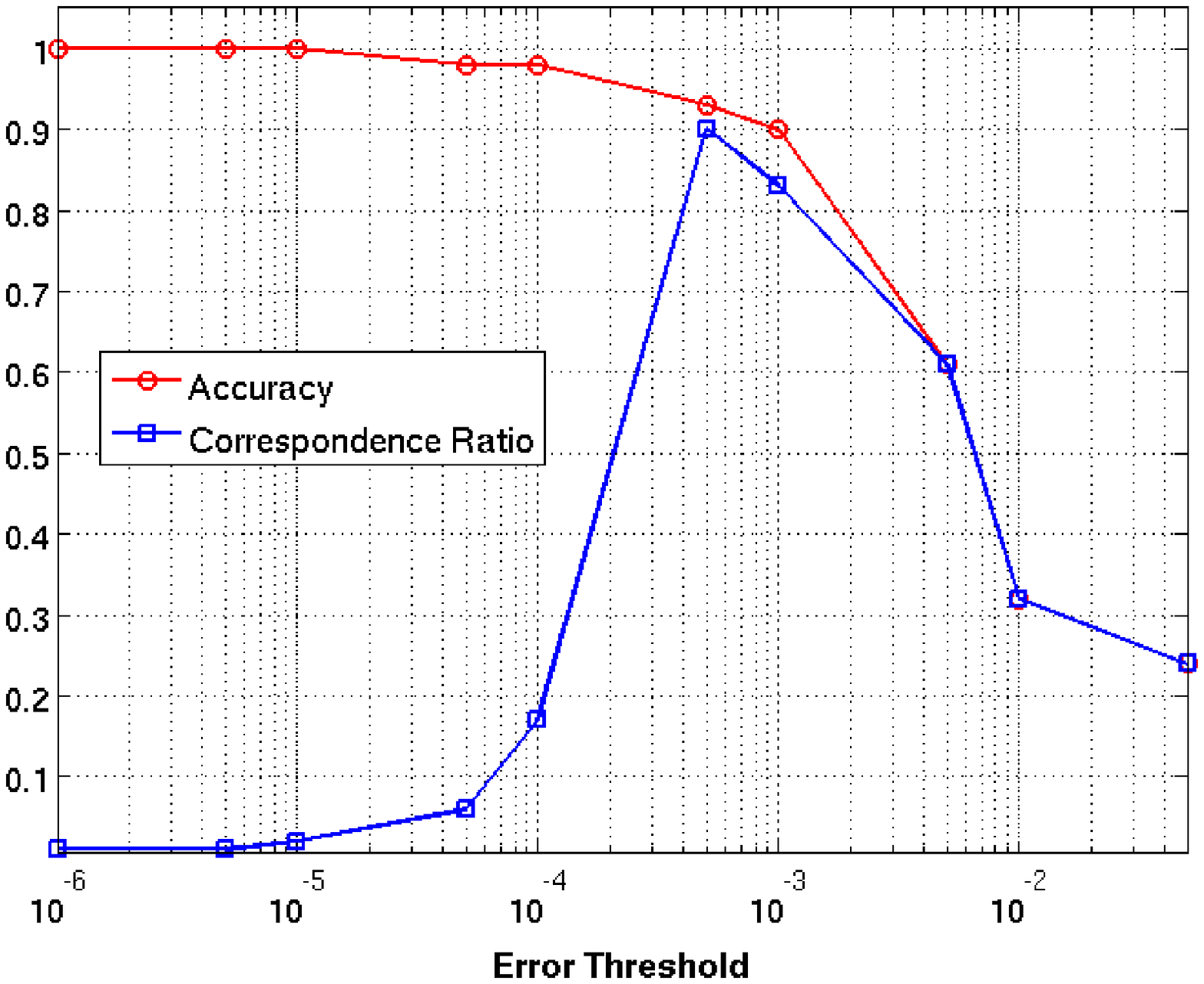,width=6cm}}
\end{minipage}
\caption{Choosing the threshold $\epsilon$ for the KSCSA segmentation algorithm. The \textit{Accuracy} and \textit{Correspondence Ratio} are plotted against different values of the error threshold $\epsilon$ for two example images. An appropriate threshold, that results in high \textit{Acc} and \textit{CR}, can be chosen using a validation dataset.}
\label{Fig:acccor}
\end{figure}

The resulting segmented images were compared to the ground truth and performance was measured using the metrics \textit{Accuracy (Acc)} and \textit{Correspondence Ratio (CR)} computed as \cite{clark1998automatic}
\begin{equation}
\label{eqn:eqn_acc}
Acc = \frac{\text{TP}}{\text{Total \# tumor pixels in the GT image}} ,
\end{equation}and
\begin{equation}
\label{eqn:eqn_cr}
CR = \frac{\text{TP} - 0.5\text{FP}}{\text{Total \# tumor pixels in the GT image}} ,
\end{equation} where TP indicates the number of true positives (the pixels indicated as tumorous by the ground truth and our algorithm), and FP denotes the number of false positives (pixels indicated as non-tumorous by the ground truth, but tumorous by our algorithm). The other unknown parameter in the KSCSA approach is the error threshold $\epsilon$, used for classifying the pixels. Figure \ref{Fig:acccor} shows the relationship between $Acc$ and $CR$ vs the error threshold ($\epsilon$) for two example images. The $\epsilon$ value was fixed at an appropriate value that resulted in high $Acc$ and $CR$ values on a validation dataset.

\begin{table}[t]
\setlength{\tabcolsep}{3pt}
\tbl{Comparison of the tumor segmentation performance obtained using (a) Active contour method (ACM), (b) Kernel sparse coding-based automated segmentation algorithm (KSCA), and (c) Kernel sparse coding-based semi-automated segmentation algorithm (KSCSA). For each patient, results for a few sample images (pre- and post-treatment) are shown. In each case, the accuracy and correspondence ratio of the segmentation in comparison to expert-marked ground truth are presented.}
{\begin{tabular}{ccccccc|ccccccc} \toprule
\small
\textbf{Image} & \multicolumn{2}{c}{\textbf{\textbf{ACM}}} & \multicolumn{2}{c}{\textbf{KSCA}}& \multicolumn{2}{c|}{\textbf{KSCSA}}&\textbf{Image} & \multicolumn{2}{c}{\textbf{\textbf{ACM}}} & \multicolumn{2}{c}{\textbf{KSCA}}& \multicolumn{2}{c}{\textbf{KSCSA}} \\
\textbf{Set}& \textbf{Acc} & \textbf{CR} & \textbf{Acc} & \textbf{CR} & \textbf{Acc} & \textbf{CR}&\textbf{Set}& \textbf{Acc} & \textbf{CR}& \textbf{Acc} & \textbf{CR}& \textbf{Acc} & \textbf{CR} \\
\colrule
\textbf{Patient 1:}&&&&&&&\textbf{Patient 6:}&&&&& \\
Pre&0.81&0.71&0.87&0.86&0.92&0.91&Pre&0.98&0.97&1&0.96&0.99&0.99\\
Pre&0.42&0.12&0.66&0.33&0.69&0.41&Pre&0.62&0.43&0.96&0.94&0.95&0.94\\
Pre&0.48&0.22&0.78&0.57&0.78&0.62&Pre&0.87&0.81&0.92&0.91&0.97&0.96\\
Pre&0.43&0.15&0.72&0.6&0.71&0.64&Post&0.91&0.87&0.92&0.87&0.93&0.91\\
Pre&0.42&0.13&0.67&0.48&0.68&0.47&Post&0.93&0.89&0.95&0.88&0.95&0.91\\
\colrule
\textbf{Patient 2:}&&&&&&&\textbf{Patient 7:}&&&&& \\
Pre&0.22&0.16&0.46&0.40&0.49&0.43&Pre&0.44&0.16&0.7&0.62&0.71&0.66\\
Pre&0.95&0.93&0.96&0.92&0.97&0.93&Pre&0.61&0.41&0.90&0.73&0.90&0.82\\
Pre&1.00&0.99&1&0.98&0.99&0.99&Pre&0.82&0.73&0.91&0.86&0.90&0.88\\
Pre&0.87&0.80&0.95&0.81&0.97&0.82&Pre&0.83&0.74&0.90&0.81&0.90&0.79\\
Pre&0.95&0.93&0.97&0.94&0.98&0.91&Pre&0.94&0.91&0.94&0.92&0.95&0.91\\
\colrule
\textbf{Patient 3:}&&&&&&&\textbf{Patient 8:}&&&&& \\
Pre&0.97&0.96&0.97&0.96&0.98&0.96&Pre&0.77&0.65&0.95&0.79&0.98&0.87\\
Pre&0.91&0.86&0.95&0.9&0.98&0.96&Pre&0.73&0.60&0.91&0.8&0.95&0.84\\
Post&1.00&1.00&0.99&0.97&1&0.99&Post&0.53&0.29&0.92&0.79&0.87&0.82\\
Post&0.76&0.64&0.98&0.81&0.97&0.85&Post&0.97&0.95&0.97&0.95&0.97&0.95\\
Post&0.81&0.71&0.83&0.73&0.86&0.72&Post&0.99&0.99&0.99&0.99&0.99&0.99\\
\colrule
\textbf{Patient 4:}&&&&&&&\textbf{Patient 9:}&&&&& \\
Pre&0.50&0.25&0.64&0.57&0.7&0.65&Pre&0.94&0.91&0.95&0.93&0.95&0.94\\
Pre&0.53&0.29&0.98&0.84&0.97&0.88&Pre&0.95&0.93&0.98&0.96&0.99&0.94\\
Pre&0.93&0.90&0.91&0.88&0.92&0.9&Post&0.47&0.21&0.87&0.75&0.88&0.78\\
Pre&0.40&0.10&0.91&0.82&0.94&0.9&Post&0.63&0.44&0.85&0.84&0.87&0.82\\
Post&0.73&0.60&0.79&0.67&0.82&0.72&Post&0.82&0.72&0.91&0.88&0.94&0.86\\
\colrule
\textbf{Patient 5:}&&&&&&&&&&&& \\
Pre&0.94&0.90&0.96&0.88&0.97&0.89\\
Pre&0.81&0.71&0.91&0.84&0.90&0.83\\
Pre&0.54&0.31&0.68&0.59&0.70&0.66\\
Pre&0.92&0.88&0.98&0.96&0.98&0.97\\
Pre&0.78&0.66&0.94&0.9&0.95&0.91\\
\botrule
\end{tabular}}
\label{Table:results}
\end{table}

\begin{figure}[t]
\begin{minipage}[b]{1\linewidth}
  \centering
  \centerline{\psfig{file=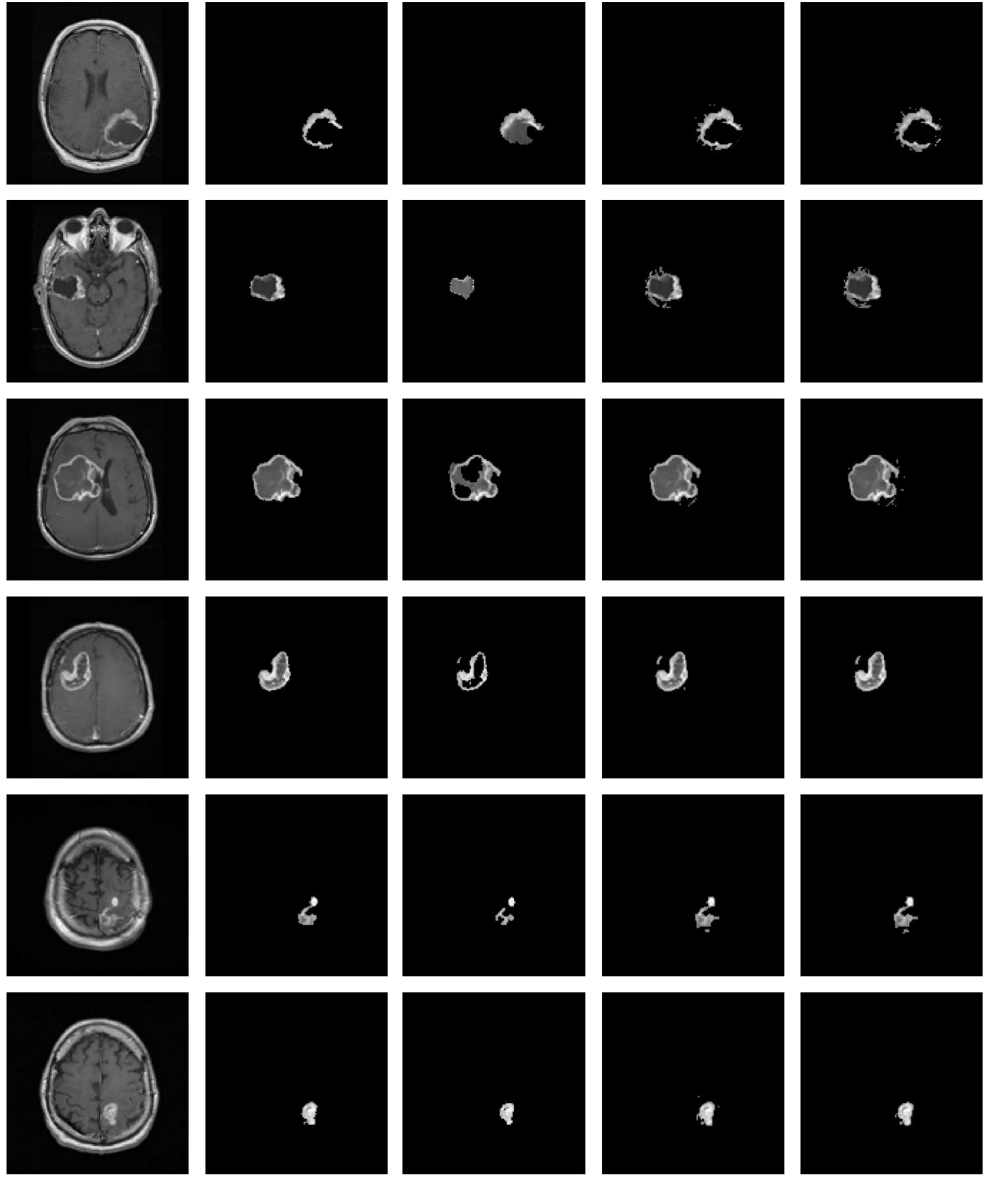,width=13cm}}
 \end{minipage}
\caption{Tumor segmentation results. (Left-Right) Original image, Ground Truth (GT) marked by an expert radiologist, Segmentation obtained using the active contour method, Segmentation obtained using the KSCA algorithm, and Segmentation obtained using the KSCSA algorithm. In all cases, the proposed algorithms provide superior quality segmentation when compared to the benchmark algorithm.}
\label{Fig:results}
\end{figure}

Figure \ref{Fig:results} shows the original and segmented images for a few example cases. In each case, the expert-marked ground truth is shown along with the results obtained using the ACM and the proposed algorithms. Both the proposed semi-automated and automated segmentation methods outperformed the benchmark method, and obtained high \textit{Acc} and \textit{CR} values as demonstrated by the extensive results in Table 1. We observed that the performance of the automated algorithm (KSCA) is equivalent to that of the semi-automated algorithm (KSCSA) in many cases and very closely comparable in the remaining cases. As expected, the semi-automated algorithm is significantly faster when compared to the automated approach. On an average, the proposed semi-automated algorithm takes about $8$ seconds (measured using MATLAB R2012a on a 2.8GHz, Intel i7 desktop) in comparison to $120$ seconds taken by the automated algorithm. Note that, the average time reported for the semi-automated algorithm does not include the time taken by the user to initialize the tumor region.

\section{Conclusions}
\label{sec:future}
A novel, automated segmentation technique for detecting brain tumors was proposed in this paper. In the new approach, we constructed ensemble kernel matrices using the pixel intensities and their spatial locations, and obtained kernel dictionaries for sparse coding pixels in a non-linear feature space. The resulting sparse codes were used to train a linear SVM classifier that determines if a pixel in the image belongs to an active tumor region. Furthermore, a semi-automated segmentation approach was proposed that uses two kernel dictionaries to model the tumor and non-tumor pixels respectively and employs a simple error-based classifier. Using simulations on a real dataset obtained for $9$ different patients, we demonstrated that both of the proposed approaches resulted in accurate tumor identifications in comparison to the widely used Chan-Vese active contour method. Future work involves extending the proposed approaches to include other types of MR imaging methods such as T2-weighted, FLAIR, perfusion-weighted, and diffusion-weighted images. Segmentation along with volumetric registration on different slices can be used to quantify the volume of the tumor region and model the growth of tumors over a period of time. The proposed algorithms can also be extended to identify tumors by computing kernel sparse codes for $3-$D volumetric data instead of $2-$D images.

\section{Acknowledgements}
We thank Dr. Mark Preul and Milad Behbahaninia from Barrow Neurological Institute for providing us with the tumor data.

\end{document}